# Parametric type design in the era of variable and color fonts

Santhosh Thottingal *
Swathanthra Malayalam Computing,
santhosh.thottingal@gmail.com

January 2025

**Abstract**

Parametric fonts are programatically defined fonts with variable parameters, pioneered by Donald Kunth with his METAFONT technology in the 1980s. While Donald Knuth's ideas in METAFONT and subsequently in METAPOST are often seen as legacy techniques from the pre-graphical user interface (GUI) era of type design, recent trends like variable fonts suggest a resurgence of certain principles. This paper explores a modern type design process built on parametric design principles, specifically using METAPOST. The author created two variable fonts with this method and released them under a free, open-source license. The paper details the methodology, workflow, and insights gained from this process.

***Keywords*** — Parametric design, Variable fonts, Opentype, Color fonts, METAFONT, METAPOST, Educational type design, Open source

## 1 Introduction

Parametric type design uses algorithms to create customized and complex typography. It involves defining parameters such as stem width, serif, and other letter form components, which can be adjusted to generate a wide range of variations within a single font. Parametric type design approach was pioneered by Donald Knuth with his METAFONT [1], technology in the early 1980s. However, this approach has not gained widespread acceptance in mainstream type design due to the perceived need for close collaboration between designers and programmers to define the design mathematically. Knuth attributes this to the fact that "asking an artist to become enough of a mathematician to understand how to write a font with 60 parameters is too much." [2].

---

*ORCID 0009-0004-7350-5234

The contemporary type design process uses advanced graphical user interface(GUI) editors. These editors not only facilitate the design aspect but also streamline the entire workflow encompassing the creation of a fully functional typeface and its subsequent proofing. It is worth noting that these GUI-based tools often come at a substantial cost, are proprietary in nature, and in some cases, are dependent on specific operating systems.

Nevertheless, certain foundational concepts introduced by Knuth have persisted and evolved over time. One such concept was the notion of generating multiple instances of a font by manipulating key parameters, a concept now recognized as variable fonts, which have garnered considerable popularity in contemporary typography [3] [4].

In light of the fact that typefaces have evolved into increasingly complex software entities rather than mere digital renditions of calligraphic designs, the discipline of type design has undergone a transformation into a branch of software engineering. This transformation has necessitated that type designers either possess a working knowledge of software engineering and mathematical principles or collaborate closely with individuals skilled in the realm of 'type engineering'.

As a professional with a background in software engineering who subsequently ventured into type design, I embarked on an endeavor to revisit METAFONT and integrate it with the modern type design workflow. The author created two variable fonts with this method and released them under a free, open-source license. The paper details the methodology, workflow, and insights gained from this process.

## 2 Parametric design

The fundamental premise of parametric design revolves around the notion that design outcomes are governed by explicitly defined parameters. Altering these parameters independently yields distinct outputs without necessitating a complete redesign. METAFONT is a programming language that helps to define such designs.

### 2.1 METAFONT and METAPOST

Donald Knuth started work on font creation software in 1977, and produced the first version of METAFONT in 1979. Due to shortcomings in the original METAFONT language, Knuth developed an entirely new METAFONT system in 1984, and it is this revised system that is used today. The AMS Euler typeface was created by Hermann Zapf with the assistance of Donald Knuth using METAFONT. Computer Modern is another typeface created using METAFONT. METAFONT produces bitmap fonts that can be embedded in PostScript documents.

METAPOST, a successor of METAPOST, is a graphic programming language developed by John Hobby that allows its user to produce high-quality graphics [5]. METAPOST produces vector graphics from a geometric/algebraic description. The language shares METAFONT's declarative syntax for manipulating lines, curves, points and geometric transformations. Unlike METAFONT,

METAPOST can produce SVG output.

```
1    beginfig(1);
2    side:=10;
3    draw (0,0) -- (side,0) -- (side,side) --(0,side) -- (0,0);
4    endfig;
```

Listing 1: METAPOST code to draw a square

Listing 1 is an example METAPOST program. It produces a square like this. ▢ Changing the value of *side* to say, 20, draws the same square with bigger size like this: ▢. Here the *side* is a parameter to the design of square. By changing the value *side*, and without any other modification to the design, we produce distinct renditions. This is the simplest explanation of parametric design. Let us try to draw the Malayalam letter Ra.

```
1    beginfig(0);
2    width:=100;  height:=100;
3
4    z0 = (x1 + width/4, 0);
5    z1 = (0, y0 + height/2);
6    z2 = (x1 + width/2, y1 + height/2)
       ;
7    z3 = (x2 + width/2, y1);
8    z4 = (x3 - width/4, y0);
9
10   pickup pencircle scaled 10;
11   draw z0..z1..z2..z3..z4;
12
13   % Add dots to indicate the points
14   pickup pencircle scaled 4;
15   for i=0 upto 4:
16       draw z[i] withcolor red;
17   endfor;
18   endfig;
19
```

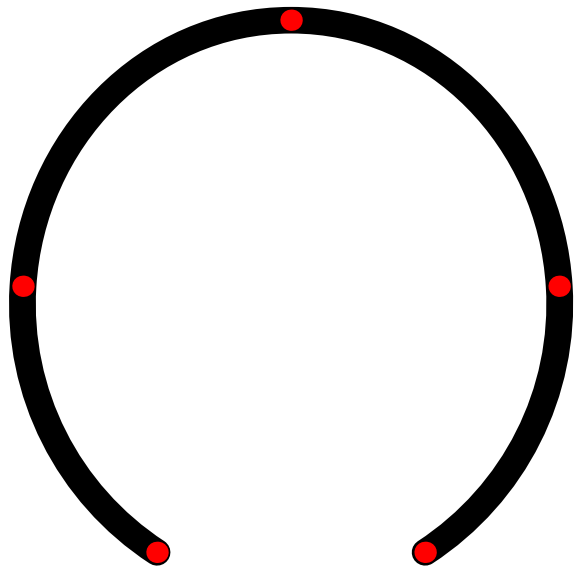

Listing 2: METAPOST code to draw the Malayalam letter Ra

From the program Listing 2, we can see that we are drawing along the 5 points $z_0$ to $z_4$ and using curves to connect them. We are using a 'pen' called pencircle scaled 10 times. Pencircle is just a circle and it will be moved along the curve and filled. There are *width* and *height* that controls the sweep of the arc too. Similar to 'pencircle', There are other predefined pens like 'pensquare'.

## 2.2 The Pens

The concept of pens, also known as nibs, are the powerful components of METAPOST. A pen could be of any shape[1]. Let us use a pencircle, but deform it first by scaling 10 in x dimension and 5 in y dimension.pickup pencircle xscaled 10 yscaled 5 makes it an elliptical pen: ⬤ Then if we move that pen along the same path, we get:

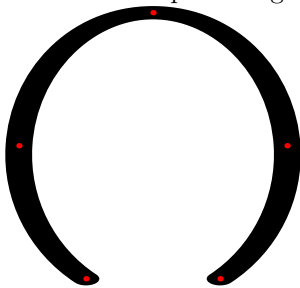

We can rotate this elliptical nib to an angle and use it as nib using pickup pencircle xscaled 10 yscaled 5 rotated 45. We get this nib: ⬤ Now, if we draw the same shape using this slanted nib, we get:

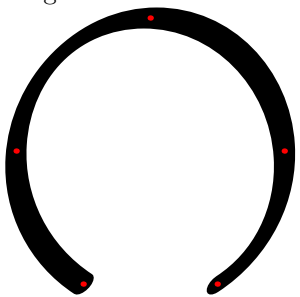

Let us attempt to define a broad nib calligraphy pen. They are straight lines without much thickness(we can assume 0 thickness, or razors), held at an angle like 40°or 45°. It will look like this: ╱ If we draw with this calligraphy pen we get:

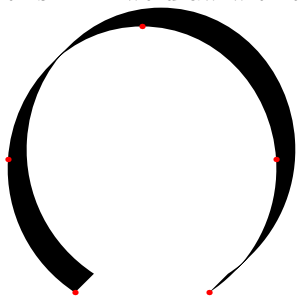

As we saw from examples above, we were able to produce a varying number of strokes from the simple pens defined. However, more complex character shapes, such as those found in serif fonts, cannot be adequately represented using pen strokes alone. One approach is to use a combination of different pens. Another approach is to draw the outlines directly instead of using the outline

---

[1]Any convex polygon, to be precise.

created using pens.

Not only the type of pen, but the aspects of pens can also be parameterized. For example, if we change the length of Calligraphic nib in previous example, we can generated the same drawing with varying stroke width:

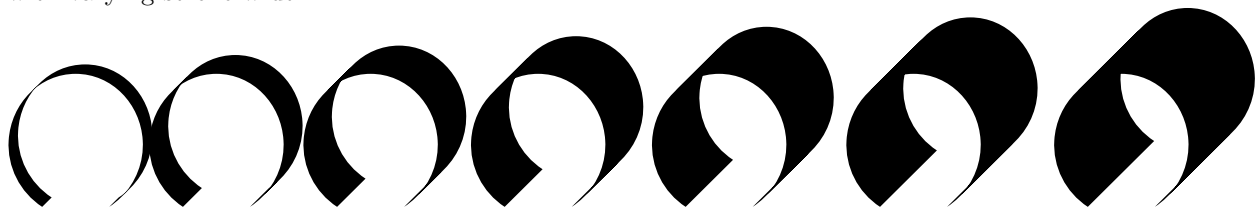

This is particularly useful in type design to generate the weight variants like thin, light, regular, bold, black etc. Since each of the drawing will have exact number of nodes, they are easily interpolatable to generate a variable font with 'weight' axis.

## 2.3 The Curves

The draw command of METAPOST connects the points using straight lines or curves. The points are defined by coordinates in the form of $(x, y)$. Points connected using dots are curves ... Points connected using – are straight lines. — connect a straight line and curve with a smooth joint. So examples:

draw (0, 0) – (20,0) –(20, 10) ⇒

draw (0, 0) .. (10,0) .. (20,0) ⇒

draw (0, 0) – (10,0) .. (20,0) ⇒

draw (0, 0) — (10,0) .. (20,0) ⇒

When METAPOST draws a smooth curve through a sequence of points, each pair of consecutive points is connected by a cubic Bézier curve, which needs, in order to be determined, two intermediate control points in addition to the end points. The points on the curved segment from points $p_0$ to $p_1$ with post control point $c_0$ and pre control point $c_1$ are determined by the formula

$p(t) = (1t)^3 p_0 + 3(1t)^2 t c_0 + 3(1t) t^2 c_1 + t^3 p_1$

where $t[0, 1]$. METAPOST automatically calculates the control points such that the segments have the same direction at the interior knots.

In the Figure below, the additional control points are drawn as green dots and connected to their parent point with green line segments. The curve moves from the starting point $p_0$ in the direction of the post control point of $p_0$, but bends after a while towards $p_1$. The further away the post control point is, the longer the curve keeps this direction. Similarly, the curve arrives at a point coming from the direction of the pre-control point. The further away the pre-control point is, the earlier the curve gets this direction. It is as if the control points pull their parent point in a certain direction and the further away a control point is, the stronger it pulls.

```
1    beginfig(1);
2    u := 2cm;
3    pair p[];
4    p0 = (0,0);  p1 = (2u,0);  p2 = (4u,0);
5    path q; q := p0{dir 90}..p1..{dir 90}p2;
6    draw q withpen pencircle scaled 2 withcolor blue;
7
8    for i=0 upto length(q):
9        dotlabel.rt("p" & decimal(i), point i of q);
10       draw point i of q
11           withpen pencircle scaled 4
12           withcolor blue;
13       p3 := precontrol i of q;
14       p4 := postcontrol i of q;
15       draw p3--p4 withcolor green;
16       draw p3 withpen pencircle scaled 4 withcolor green;
17       draw p4 withpen pencircle scaled 4 withcolor green;
18    endfor;
19    endfig
20
```

By default in METAPOST, the incoming and outgoing direction at a point on the curve are the same so that the curve is smooth. The algorithm behind the curve smoothening in META-POST is developed by John D. Hobby. [6][2]. METAPOST allows the control points to be specified directly in the following format:

draw (0,0)..controls (26.8,-1.8) and (51.4,14.6)..(60,40)

---

[2]The hobby package with Tikz package of LaTeXcan also create similar smooth curves. Before developing the METAPOST based workflow, I had attempted creating SVGs using Tikz, however abandoned because of various reasons such as inflexible, SVG issues. https://ctan.math.illinois.edu/graphics/pgf/contrib/hobby/hobby.pdf

It is easy to adjust these control points in a GUI editor, but in METAPOST, we want to think about only the curves. For this, METAPOST provides a mechanism to declare the direction, tension and curl of the curves along with path expression.

draw (0, 0){dir 30}..{dir 45}(10, 10)..{right}(20, 0)..{dir 30}(30, 10) = 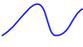

To illustrate how the direction specification changes the curves, let us plot curves with varying directions as below:

```
1  beginfig(0)
2  for a=0 upto 9:
3      draw (0,0){dir 10a}
4      ..{dir -10a}(4cm, 0)
5      ..{dir 10a}(8cm, 0) withcolor blue
       ;
6  endfor
7  endfig;
```

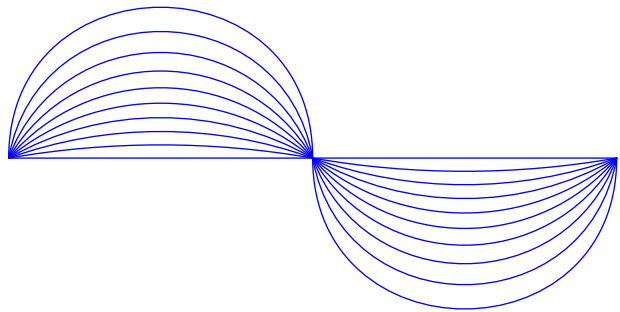

METAPOST also allows to specify the tension and curl of the curves. Since they are not used in the typefaces I designed, and for brevity, we are skipping the explanation here.

## 2.4 SVGs from METAPOST

Latest version of METAPOST can output SVGs. The previous examples of rendering a glyph with different pens might make us believe that we can directly use those SVGs for using it in a font. However, a close inspection of the SVGs produced will disappoint us. Let us revisit the 'Ra' example:

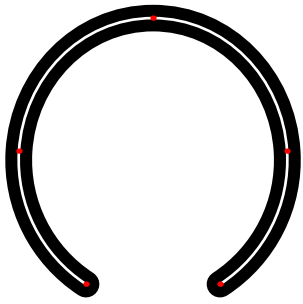

(a) SVG output from pencircle method

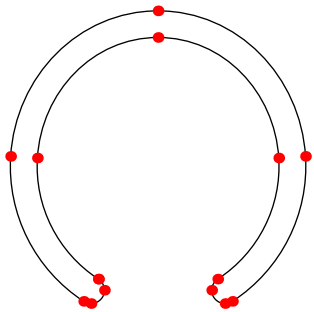

(b) Expected SVG for use in a typeface

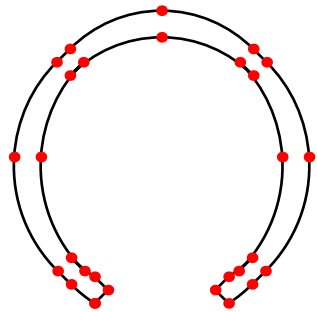

(c) Envelope(outline stroke) calculated from a convex polygon pen

Figure 1: Comparison of SVGs produced by METAPOST

In Figure 1a the white line is the path we defined and drew. The thick black line along that path is just the stroke width of the path. So this is an SVG with just 5 nodes. This is not an SVG we can use for type design. In a glyph, we need a closed outline. Our expected drawing should be as shown in Figure 1b.

While designing fonts with METAFONT Knuth outlines [7] a method for 'envelope' calculation. Internally, in METAFONT all the pencircle are approximated a convex polygon with 24 sides. Then this polygon is used just like pensquare(that has 4 sides). Then the outlines of the polygon pen is calculated to get the 'envelope' or outline stroke. This works well for the METAFONT since it outputs bitmaps. But for METAPOST since the output is a vector image, and since we want the SVG a very clean vector with very minimal nodes, this solution has issues as we see in as shown in Figure 1c. It produces several unwanted nodes and they are unpredictable when the parameters like width/height etc changed. We resolved this issues by using the macros provided by MetaType package.

## 2.5 MetaType

MetaType is a tool created by Bogusław Jackowski, Janusz Nowacki, and Piotr Strzelczyk for creating PostScript Type 1[3] fonts [8]. MetaType was used to create the Latin Modern fonts, derived from Computer Modern fonts but including many more accented characters [9]. Most important fonts produced with MetaType1 are: Latin Modern, Latin Modern Math, TeX Gyre, Antykwa Toruńska, Antykwa Półtawskiego, Kurier and Iwona.

Even though we are not generating any PostScript fonts here, MetaType was very crucial in the development of workflow. The utility library as part of MetaType include many macros to produce accurate pen envelopes suitable for typefaces.

Nupuram project extensively used macro *pen_stroke* defined in MetaType. Macro *pen_stroke* performs an operation known as "expanding stroke"; we'll call the result of the operation a "pen envelope" (for a given path). The macro has one optional parameter, *opts* ($text$), and two obligatory ones: input path $p$ ($expr$) and a $result$ ($suffix$). A user has an access to subpaths of the envelope, namely: $result_r$ is the right edge of the envelope, $result_l$—its left edge, $result_b$—is a fragment of the pen outline joining left and right edge of the envelope at the beginning node of the path, $result_e$—is a similar fragment at the ending node of the path (see the picture below). If the path $p$ is cyclic, then $result_e$ and $result_b$ are undefined, otherwise the variable $result$ contains additionally the complete expanded stroke.

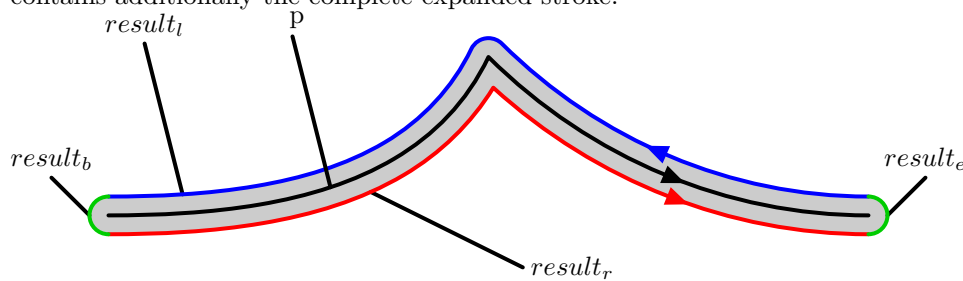

For finding an envelope, a default path ($default\_nib$, returned by $fix\_nib$) is used except nodes for which the parameter *opts* sets another pen. Mastering the usage of the parameter *opts* allows a user to achieve nontrivial effects. The parameter *opts* is a list (space-separated or semicolon-separated) of the following operators: (1) *nib*, (2) *cut*, (3) *tip*, and (4) *ignore_directions*.

The macro *nib* has two parameters: *nib*(pen)(list_of_nodes), where 'pen' is a path returned by macro $fix\_nib$, and 'list_of_nodes' contains comma-separated numbers (times) of the nodes of the path $p$ at which a given pen is to be used.

To illustrate this, let us revisit the same arc examples. As shown below, by using the *pen_stroke* macro and using different pens at different nodes in the path, we get a perfect vector output. We can see the the arc is drawn using 5 points. And corresponding to each each node, 2 nodes are present in the outline stroke, on outer and inner side of outline stroke. Depending on the pen thickness at each node, we get smooth curves that varies the thickness.

---

[3]Type 1 fonts are a legacy format created by Adobe in 1984

The resulting stroke modulation, its extrapolatable nature with respect to the parameters like nib thickness are the key tools to build a typeface, potentially a variable typeface. The terminals can be a cut at any angle or any nib at any angle or width. Each nodes in the path can be any nib at any width, shape and angle.

```
1    input plain_ex;
2    beginfig(0);
3    width:=200;    height:=200;
4    thick:=width/5;   thin:=thick*2/3;
5    % .. path definition goes here ..
6    path p, s;   p := z0..z1..z2..z3..z
     4;
7
8    vardef thicknib = fix_nib(thick,
     0, 0) enddef;
9    vardef thinnib = fix_nib(thin, 0,
     0) enddef;
10   pen_stroke(
11       cut(thinnib, 45)(0)
12       nib(thinnib scaled 1.2 rotated
     −10)(1)
13       nib(thicknib rotated 80)(2)
14       nib(thicknib rotated 10)(3)
15       cut(thinnib scaled 1.25, rel
     90)(4)
16   )(p)(s);
17   draw s withpen pencircle scaled 1;
18   fill s withcolor .8white;
19   endfig;
20
```

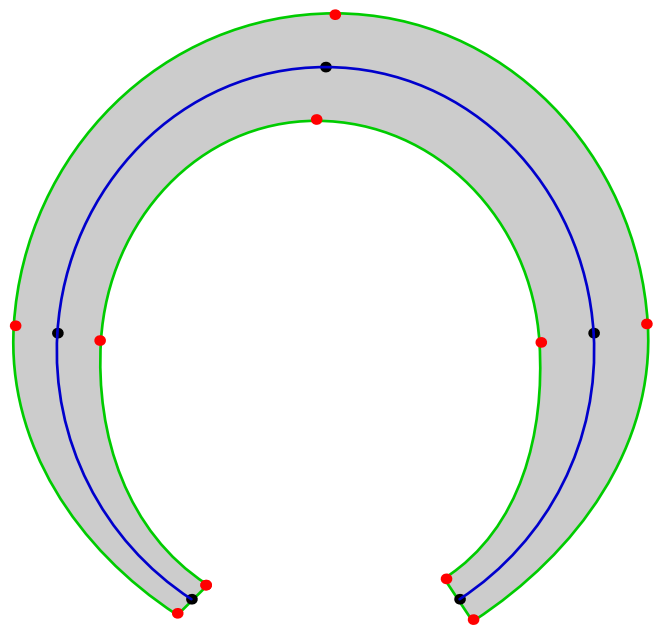

Listing 3: Usage of *pen_stroke* macro.

## 3   Earlier attempts

Although METAFONT never became mainstream, hundreds of fonts have been created with it: an incredible list has been compiled by Luc Devroye[4].

Jeroen Hellingman created Malayalam metafonts in 1994 [10]. These were created as uniform strokes. Karel Piska later converted all these fonts to Type1 fonts. He did this by an accurate analytic conversion to outlines using METAPOST output. After theoretical conversion, the Font-Forge is used for removing overlap, simplification, rounding to integer, auto hinting, generating outline fonts, and necessary manual modifications.[5]

---

[4]List of fonts created using METAFONT. Compiled by Luc Devroye. http://luc.devroye.org/metafont.html

# 4 Modern typeface design workflow with METAPOST

As we explained earlier, the first step is to define the glyphs using METAPOST. Then we compile them to SVGs. These SVGs are then converted to glif format of Unified Font Object format[6]. The Unified Font Object (UFO) is a cross-platform, cross-application, human readable, future proof format for storing font data. This is the modern format used by typ edesigners. Type design tools offer a feature to export their internal format to UFO too[7].

Once we have UFO formatted font source, we need to write the Opentype features and save it as part of same UFO file. Then type compiling tools like fontmake[8] can compile this UFO to a font binary such as .ttf, .otf or webfonts. The opentype features, glyph to unicode mapping, kerning and font meta information - these are all part of the prepared UFO. A set of programs prepares all of these based on a simple configuration file.

Preparing the opentype features used to be a major task for typeface design for Malayalam. We used to manually write them. In Nupuram typeface they are automatically generated based on the script grammar encoded in application logic.[9]. For a script Malayalam, Opentype rules are integral part of typeface. Rapid experimentation and variation generation with Nupuram was possible because of auto-generation of opentype rules.

---

[5]CTAN: indic-type1 – Indic Type 1 fonts converted from public METAFONT sources https://ctan.org/pkg/indic-type1

[6]https://unifiedfontobject.org/versions/ufo3/

[7]For example, Glyphs application's .glyphs format can be converted to ufo format.

[8]fontmake : https://github.com/googlefonts/fontmake

[9]Source code repository of Nupuram https://gitlab.com/smc/fonts/Nupuram

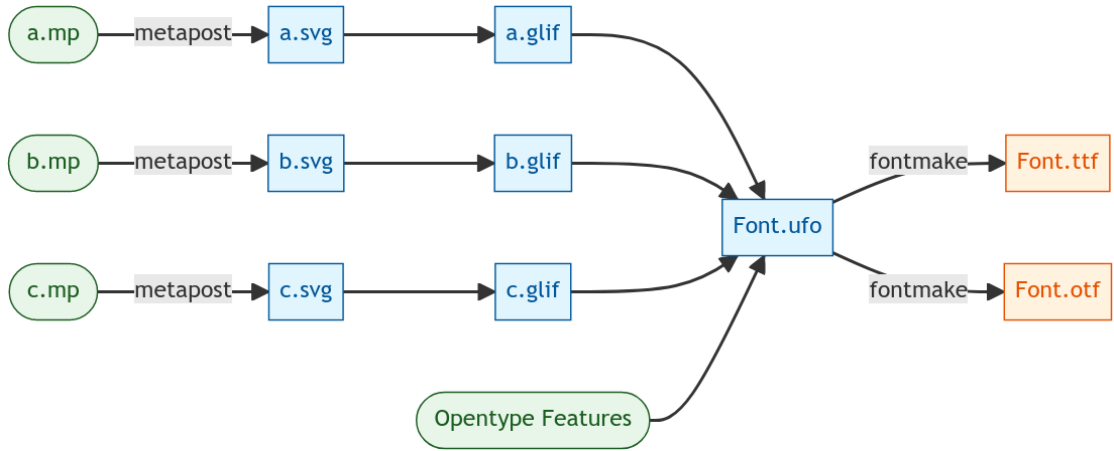

Figure 2: High level workflow of type design with METAPOST

# 5 Nupuram Type Family

## 5.1 Design

Nupuram[10] draws the inspiration from the title posters of early Malayalam movies around 1960-1970, particularly those crafted by the renowned title designer, S Appukkuttan Nair(Popularly known as SA Nair)[11]. These title designs features letterforms with wide, flat, sharp terminals, thin vertical strokes and thick horizontal strokes.[12]. Even though there are hundreds of posters done by SA Nair with same design concept, there is no strict uniformity in these designs due to their entirely handmade nature. Adapting these distinctive traits to a typeface required many customization. In this endeavor, while I departed from the pronounced sharpness of the terminals, I preserved the broad strokes while slightly reducing their thickness.

The vertical stems in glyphs are thin and horizontal ones are thick, akin to the characteristics of reverse contrast typefaces. The letters are close to handwritten style than a regular print style. One can easily notice the playful, casual, personal aesthetic. Usable at display and text sizes too.

Based on the macros we explained in Section 2.5, the nibs of Nupuram glyphs are defined. There are three 3 nibs - *thicknib*, *thinnib* and *terminalnib*. Here, *thicknib*, *thinnib* are razors nibs or nibs with 0 width. Their lengths are parameterized. *terminalnib* is a razor nib or an elliptical nib based on whether terminal are rounded(soft) or flat(sharp).

---

[10]The word Nupuram means 'anklet' https://en.wikipedia.org/wiki/Anklet

[11]SA Nair, Malayalam Movie and Music Database https://m3db.com/sa-nair

[12]Examples: Thakilu Kottampuram(https://m3db.com/film/thakilu-kottampuram), Vilakkum Velichavum (https://m3db.com/film/vilakkum-velichavum), Angadi(https://m3db.com/film/angadi)

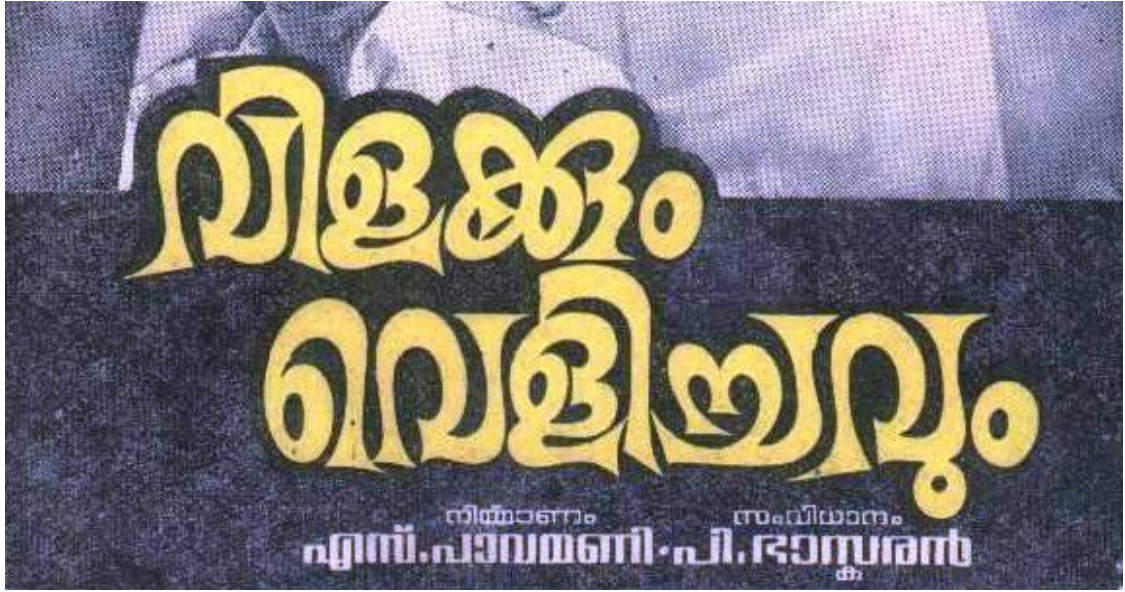

Figure 3: Vilakkum Velichavum title design by SA Nair.

 നുപുരം മലയാളം

Figure 4: Nupuram design

## 5.2 Typographic units

The nib length(that results the thickness of strokes) is defined by *thick* and *thin* configuration. The value *thick* is expressed in relationship with EM Size. the emsize is defined as $em := 1000$. Then this Em value is divided to basic typography units, defined as $u := em/10$. The $u$ is the unit we will use to define all typographic parameters such as ascent, descent, bearing, thick, thin etc.

```
1    em :=          1000;         % Height of characters - Em square
2    u :=           em/10;        % Unit width. Em square divisions.
3    ascent :=      8u;           % Ascender Height
4    descent :=     2u;           % Descender Height
5    xheight :=     2/3*ascent;   % Height of English small letters
6    mheight :=     3/4*ascent;   % Height of Malayalam letters
7    Xheight :=     8u;           % Height of English capital letters
8    thick :=       1u;           % Thickness of thickest lines
9    thin :=        0.7;          % Thickness of thinnest lines-ratio of
                                    thick.
10   subthick :=    0.666u;       % Thickness of thickest lines in subscribed
                                    characters
11   xthick :=      1;            % Extra thickness for terminals
12   slant :=       0;            % Slant of characters.Give angle values
13   condense :=    1;            % Condense factor. < 1 for condense, > 1
                                    for expand
14   lbearing :=    0.4u;         % Default left bearing
15   rbearing :=    0.4u;         % Default right bearing
16
```

Listing 4: Basic typographic parameters defined in Nupuram typeface

By changing any of these base parameters we can build a new design. But not all changes produce a good or useful design. To make a typeface with weight variants, we change the *thick* parameter alone. To change the extra thickness of terminals we change the *xthick* parameter. To get a slanted glyph with 15°, we give $slant = 15$ and so on. We will discuss this in detail in next section about variable fonts.

Using all these parameters let us try to draw the letter 'Ra', but this time as per the Nupuram design. See Listing 5

```
1  input plain_ex;
2  beginfig(0);
3  width:=200; height:=200; thick:=width/5;
4  thin:=0.7; m:=width;   xthick:=1.1;
5  terminalround:=0.5;
6  z0=(x1 + m/4 , 0);   z1=(0, m/2);
7  z2=(x0 + m/3, y1 + m/2);   z3=(x2 + m/3, y2–m
       /2);
8  z4=(x2, 0);
9  path p,s; p:=z0{dir 135}..z1.. z2{right}..z3{
       dir 260}..z4;
10
11 vardef thicknib = fix_nib(thick, thick, 0)
       enddef;
12 vardef thinnib = fix_nib(thick*thin, thick*
       thin, 0) enddef;
13 vardef terminalnib = fix_nib(thick*xthick,
       thick*terminalround, 0) xyscaled(xthick,
       terminalround) enddef;
14 vardef terminalangle expr t of p = angle(
       direction t of p)+90 enddef;
15 pen_stroke(
16     nib(thinnib)(1,3) nib(thicknib)(2)
17     nib(terminalnib rotated terminalangle 0 of
        p)(0)
18     nib(terminalnib rotated terminalangle 4 of
        p)(4)
19 )(p)(s);
20 draw s; fill s;
21 endfig;
22
```

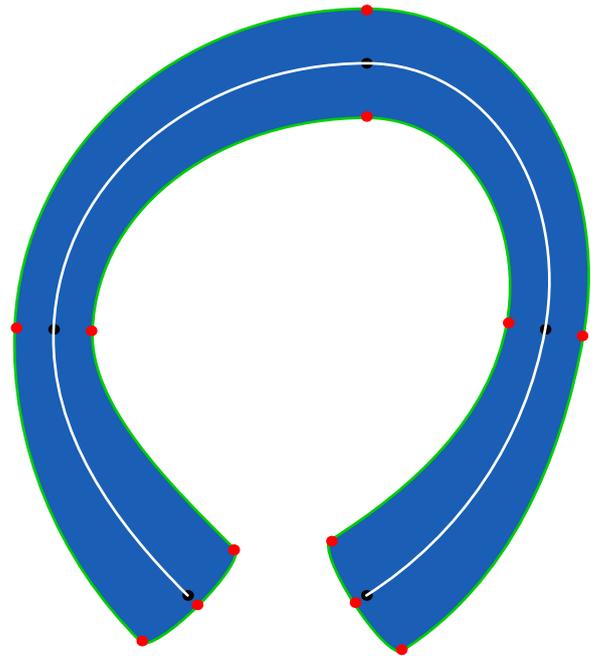

Listing 5: The 'Ra' letter in Nupuram.

### 5.3 Variable fonts

The glyphs and their variations are fully controllable by simple configurations files. On top of this common configurations, we define font specific variations in simple configuration files. For example, the configuration for creating Nupuram-Bold font will look like this

```
1    input ./config/Regular;
2    thick:= 1.25u;
3
```
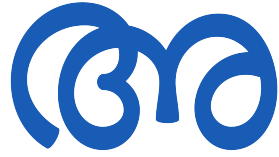

It imports(includes) the Regular configuration and set *thick* = 1.25*u*. You will quickly see that it is an increase on the *thick* value set in the configuration for 'Regular' variant

```
1    thick := 0.90u;
2    soften := 0;
3
```
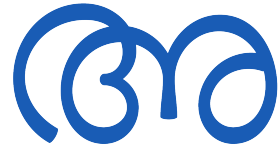

Let us look at configuration for Nupuram Thin variant by setting *thick* = 0.5*u*.

```
1    input ./config/Regular;
2    thick  :=  0.5u;
3
```
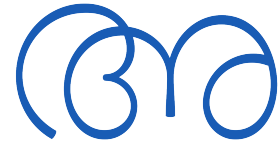

Similarly, the configuration for Nupuram Condensed is defined by setting *condense* := 0.8

```
1    input ./config/Regular;
2    condense :=  0.8;
3
```
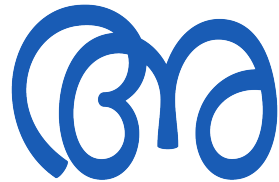

An oblique variant will have the following configuration *slant* := 15

```
1    input ./config/Regular;
2    slant :=15; % Degree of slanting
3
```
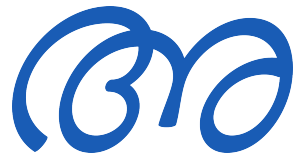

Whether the terminals are flat(sharp edge) or rounded(smooth) is controlled by the *terminalnib* that uses *terminalround* parameters. By setting it *terminalround* := 0.15 we can get sharp terminal.[13]

---

[13]Note that we are not setting it to 0 because, at zero the curve segment at terminal becomes a straight line. This affects the interpolation. We can only do interpolation of curve with another curve with more or less curl

Table 1: Nupuram variation axis tags, their ranges, default, and descriptions

| Axis | Tag | Range | Default | Description |
|---|---|---|---|---|
| Weight | wght | 100 to 900 | 400 | Thin to Black. Can be defined with font-weight CSS property. |
| Slant | slnt | -15 to 0 | 0 | Upright (0°) to Slanted (about 15°). |
| Width | wdth | 75 to 125 | 100 | Condensed to Expanded. Can be defined with usual font-stretch CSS property. |
| Soft | SOFT | 0 to 100 | 50 | Sharp to normal to SuperSoft terminals. |

```
1    input ./config/Regular;
2    terminalround:=0.15;
3
```

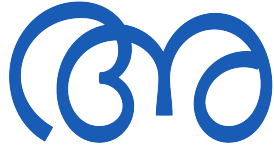

Since all of the above variants are same design, with exact number of SVG nodes and same curves, they are interpolatable. We can use them as master designs in a variable fonts designspace and get a variable font with 'weight', 'width', 'slant' and 'soft' axis.

All the 4 axes can be considered as a 4 dimensional space. By carefully choosing the values of any of these 4 axis, we can practically produce infinite styles. Modern tyepsetting softwares, Web pages(CSS) allow selecting this style by quick previews. Even though a user can choose any of these styles, the typeface also comes with a predefined 'named instances' which are predefined combination of these values. For example, 'Nupuram Condensed Thin' is a named instance that choose 'width' and 'weight' axis value as 'Condensed' and 'Thin'. Nupuram has 32 such named instances.

Nupuram is the first variable font in Malayalam with 4 design axes.

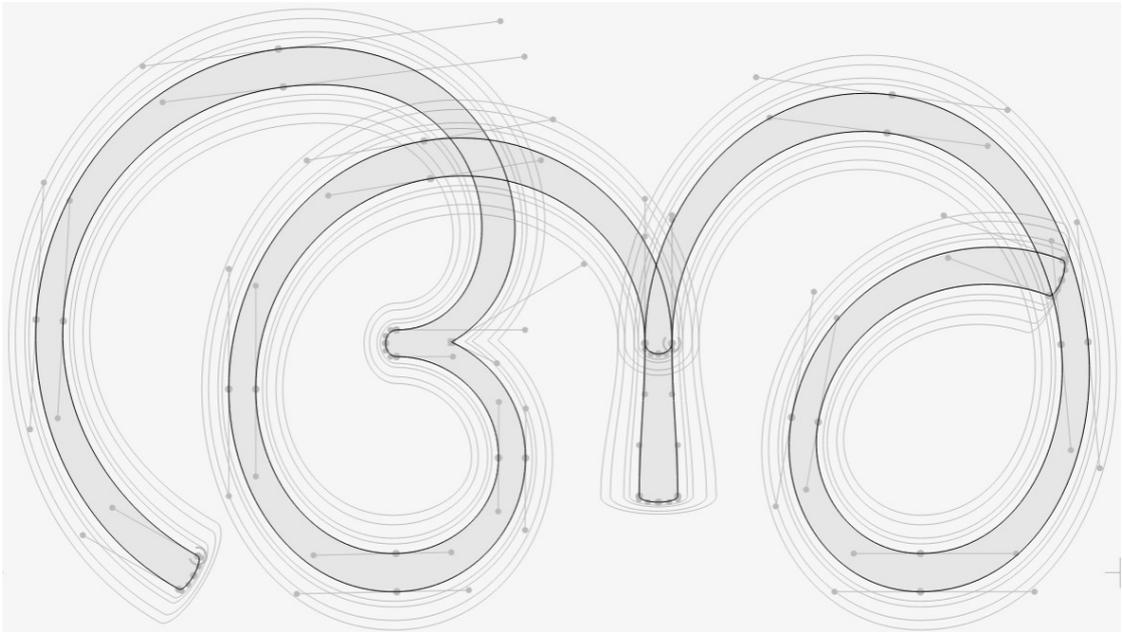

Figure 5: Visualization of Nupuram wght axis. The thickness of strokes varies here.

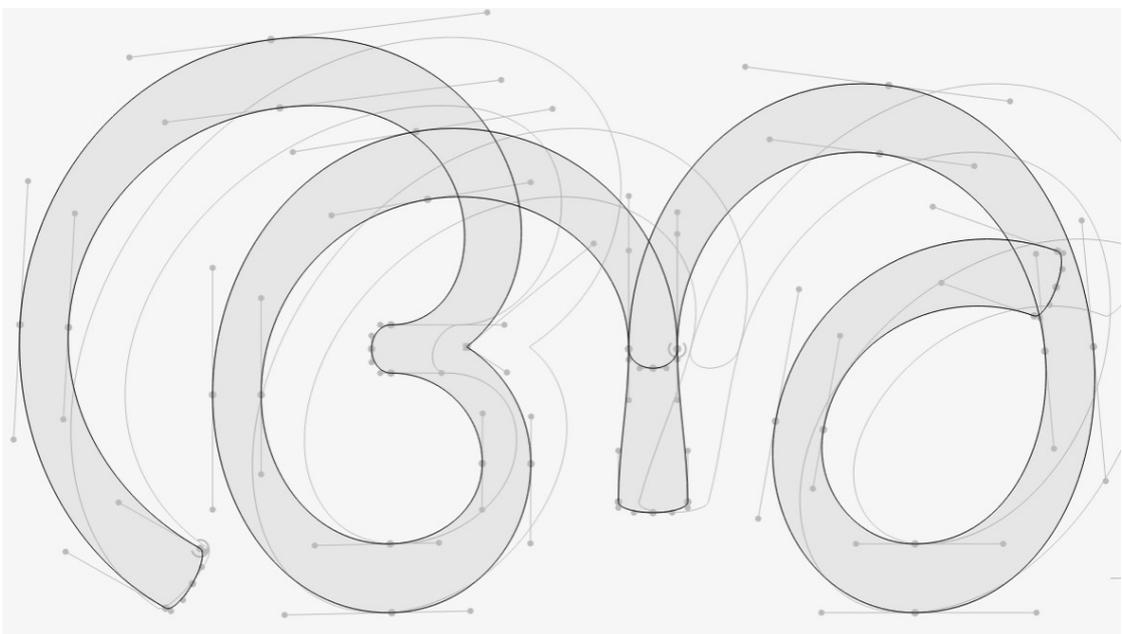

Figure 6: Visualization of Nupuram slnt axis. The slant angle changes from 0° to -15°.

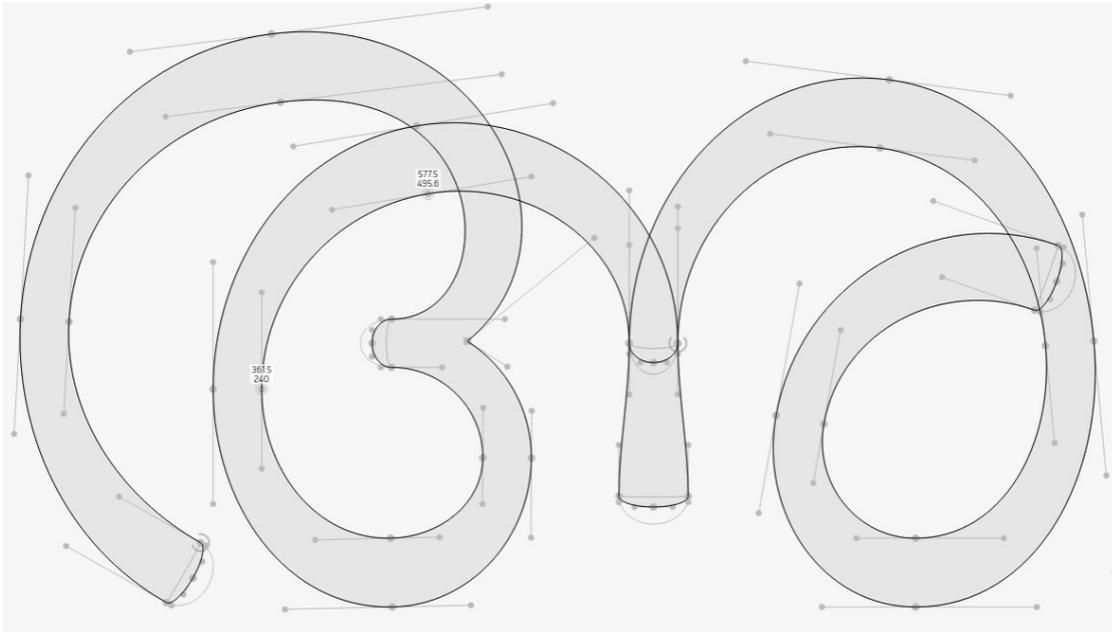

Figure 7: Visualization of Nupuram SOFT axis. The sharpness or softness varies from 0 to 100

## 5.4 Debugging and proofing

GUI based font editors usually has advanced preview and proofing systems integrated. This is also an essential requirement for any type design system. While designing Nupuram, I used Visual Studio programming IDE along with live preview of SVGs. Figure 8 shows an example setup. In development mode, on top of the glyphs, guidelines, coordinates, points are also visualized to help the design process.

However, we also need more advanced proofing system to render an arbitrary text content with the given font to evaluate and fine tune various design aspects. For this, I developed a webbased type test and preview system[14]. See Figure 9

To ensure the quality of glyphs and various technical details to be taken care for a production ready typeface, automatic testing is incorporated using fontbakery tool[15].

---

[14]This online tool is available at https://smc.gitlab.io/fonts/Nupuram/tests/
[15]fontbakery: A font quality assurance tool https://github.com/fonttools/fontbakery

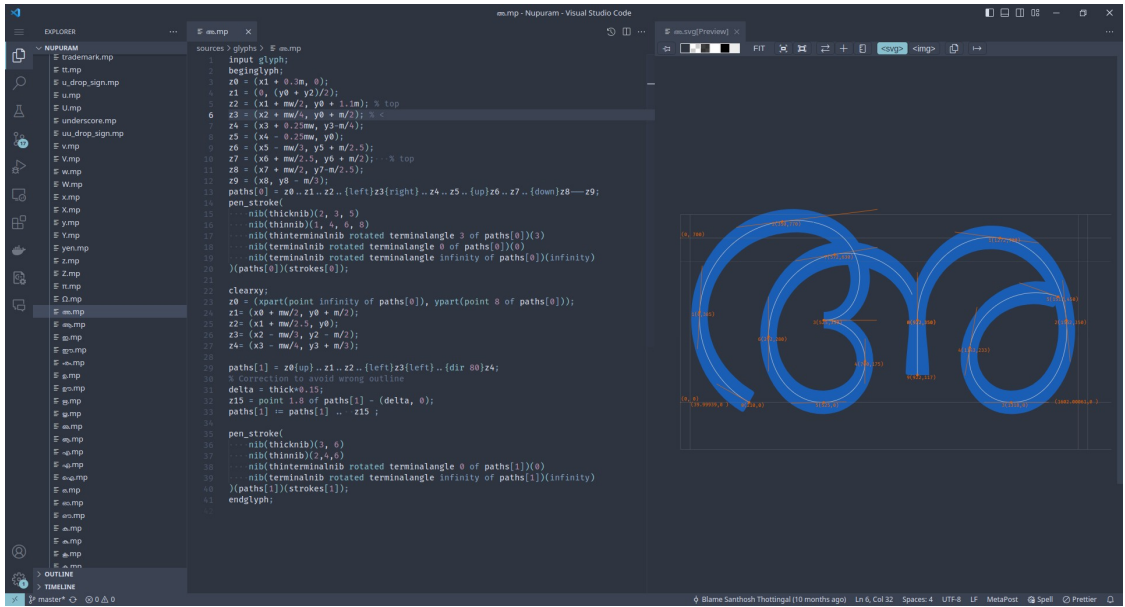

Figure 8: An example typedesign setup with VS Code IDE with METAPOST and SVG generated shown side by side. Changing the code automatically refreshes the image.

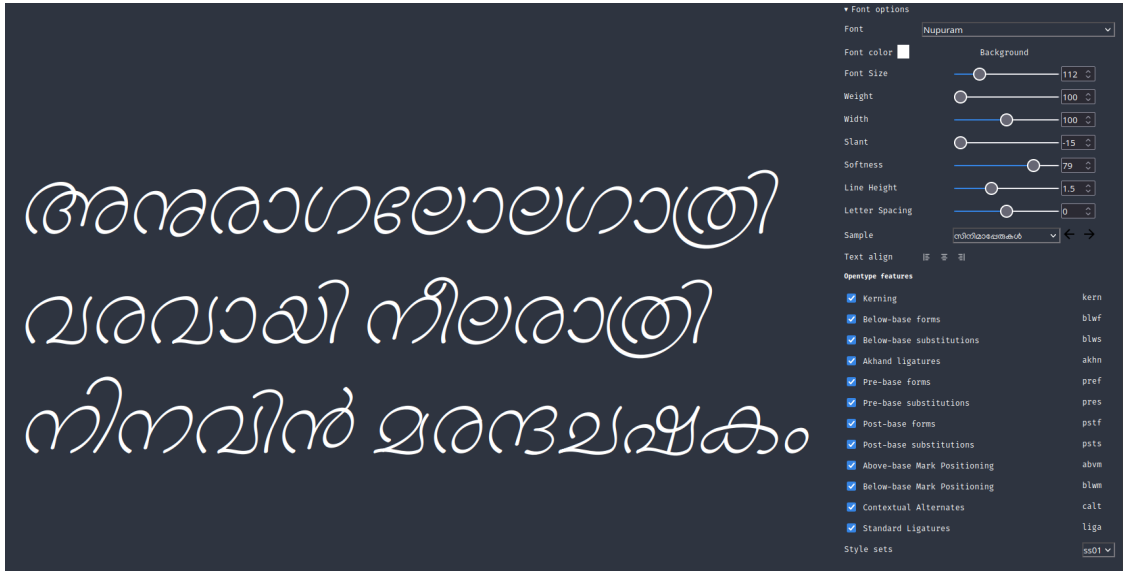

Figure 9: web based type testing, proofing, playground used in the Nupuram design process

## 5.5 Latin glyphs

All the Malayalam characters defined in Unicode version 15 are present in the font. Nupuram also has latin script support. Nupuram supports 294 languages covering approximately 2.8B speakers[16]

The latin glyphs follow the same design of Malayalam. See Figure 10

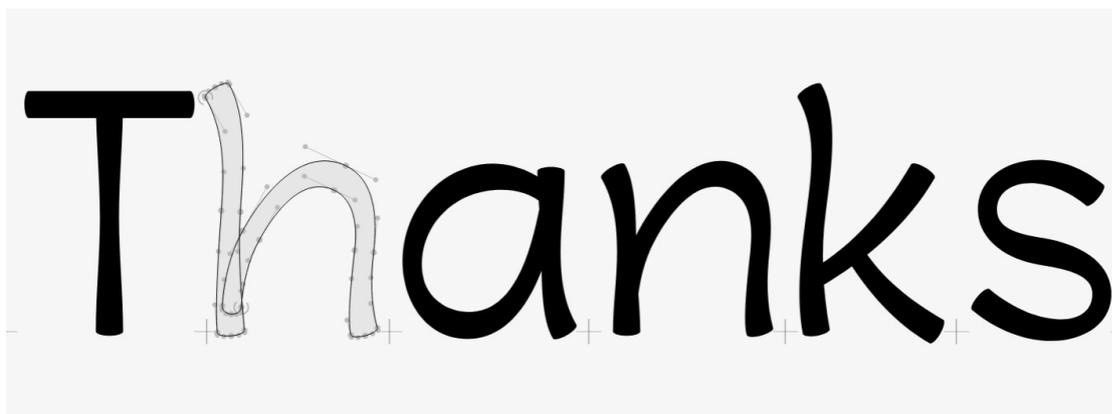

Figure 10: Nupuram English sample

The Earth is a very small stage in a vast cosmic arena. Think of the rivers of blood spilled by all those generals and emperors so that, in glory and triumph, they could become momentary masters of a fraction of a dot. Think of the endless visited by the inhabitants of one corner of this pixel the scarcely distinguishable inhabitants of some other corner, how frequent their misunderstandings, how eager they are to kill one another, how fervent their hatreds. The Earth is the only world known so far to harbor life. There is nowhere else, at least in the near future, to which our species could migrate.

Figure 11: Nupuram sample English paragraph

## 5.6 Nupuram Calligraphy

Nupuram Calligraphy simulates a wide nib Calligraphy pen with nib rotation at 40°. This is a variable font with 'weight' axis. The width of the calligraphy pen can be varied for getting different weights. The width of the nib can be varied as required. So by defining 3 widths, narrow, medium, wide, we can get 3 variants of this glyph. Using these master glyphs, we can

---

[16]Calculated using hyperglot tool https://hyperglot.rosettatype.com/

interpolate to any nib size using the variable font technology. That is how we made Nupurum Calligraphy variable font. An illustration of weight axis variations is given in Figure 12

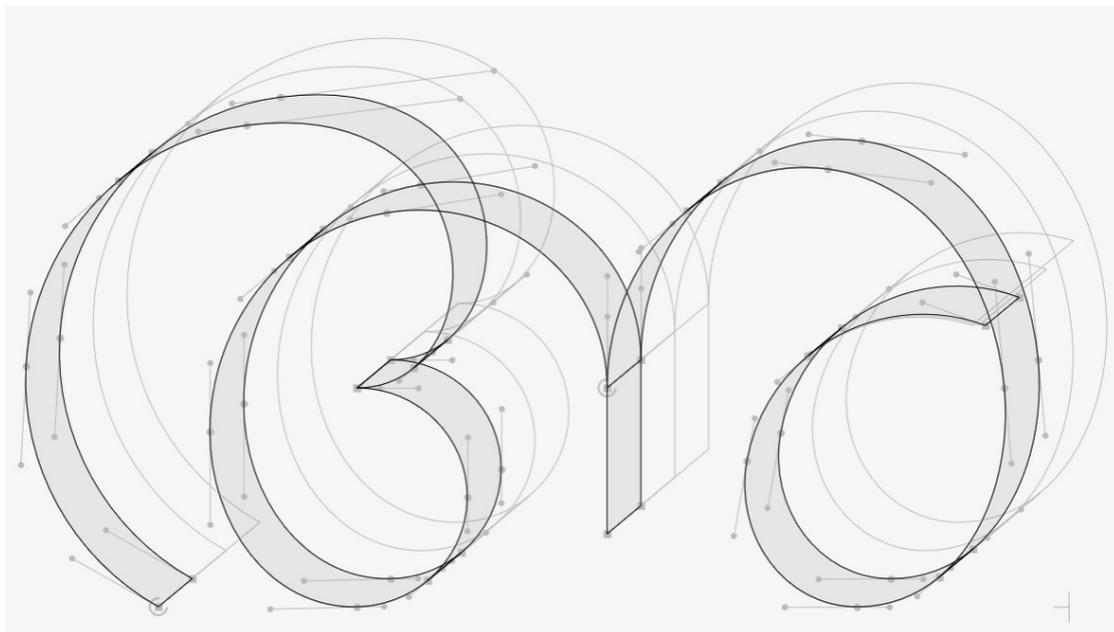

Figure 12: Weight axis variations of Nupuram Calligraphy subfamily

അലകടലിനക്കരെ

Figure 13: Nupuram Calligraphy sample

## 5.7 Nupuram Color

Color fonts (also known as chromatic fonts) can use multiple colors, including gradients, in a single glyph, rather than the flat, single color used by typical, non-color (monochromatic) fonts. This relatively new technology allows designers to set the color palette within the font to express themselves with color in a way that would previously not be possible outside of advanced graphics applications. [17]. Nupuram has a Color font version with COLRv1 specification. The colors can be customized by users, for example using CSS. Nupuram Color is also a variable font with customizable 'wght' axis.

---

[17]Introducing color fonts, Rod Sheeter https://fonts.google.com/knowledge/introducing_type/introducing_color_fonts

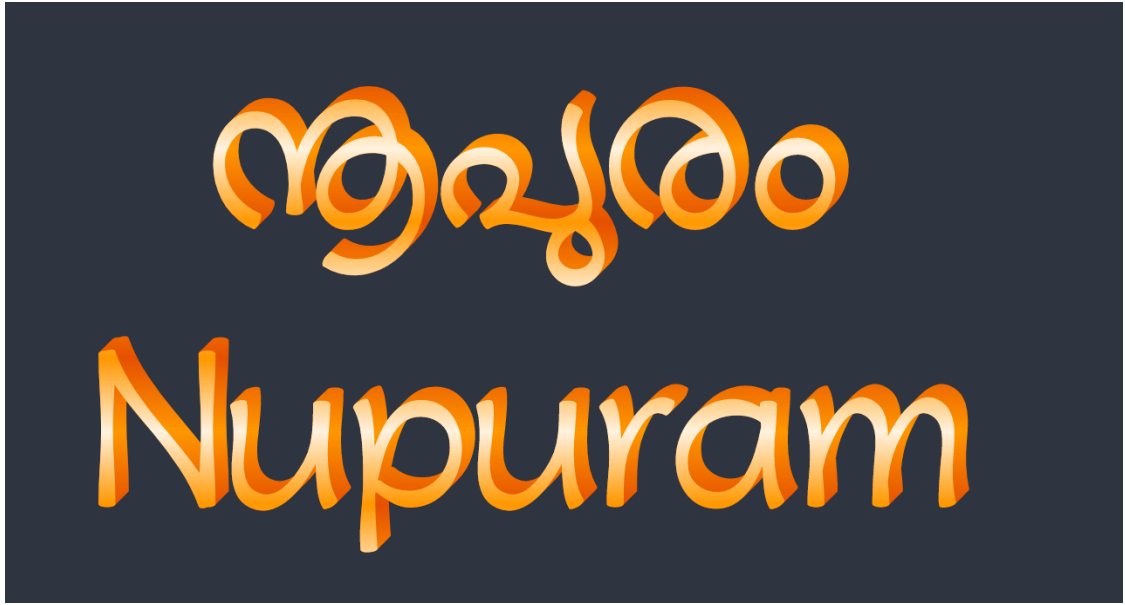

Figure 14: Sample rendering of Nupuram color font with default color palette

Nupuram Color creates a solid 3D objects illusion. The anatomy of color font is based on an observation in Nupuram Calligraphy. In the calligraphy variant, we moved a calligraphic pen through the glyph path. If we use pen strokes with thick and thin strokes as explained earlier, we get a modulated glyph outline as Figure 15a. If the calligraphic pen we used for Nupuram Calligraphy is move through the outline, we will get Figure 15b

It is slightly confusing drawing, but we start to see a 3D shape in it. Let us fill this with color blue to get a better picture. See Figure 15c. Now we can relate this with the calligraphy glyph we constructed earlier. Same stroke modulation, but two times - for outer and inner lines.

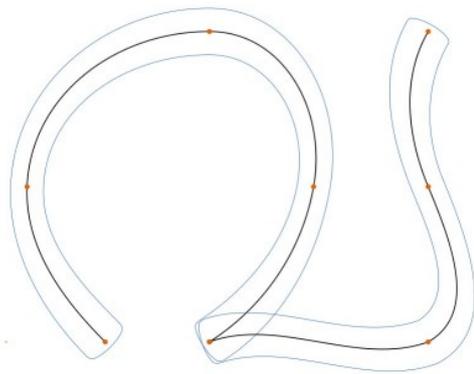

(a) Letter 'va' with pen stroke and outlines.

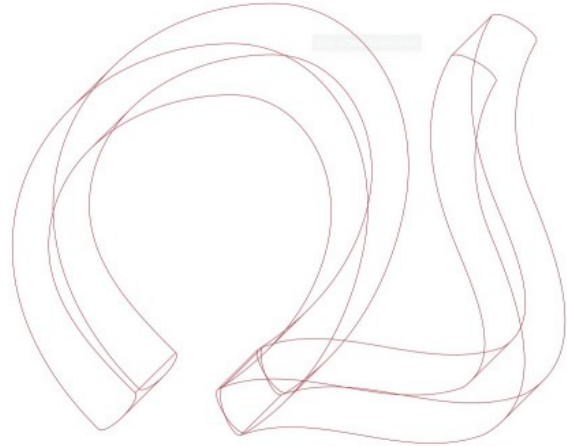

(b) Moving calligraphy nib along the outline

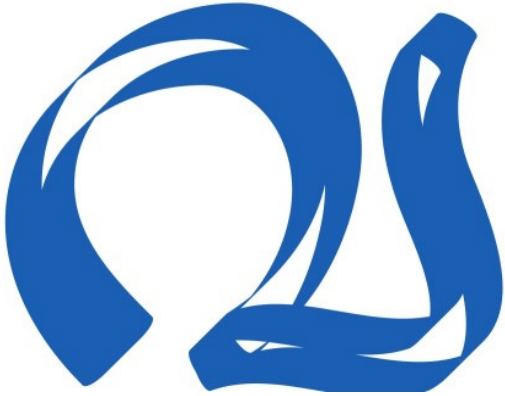

(c) Figure 15c filled with color.

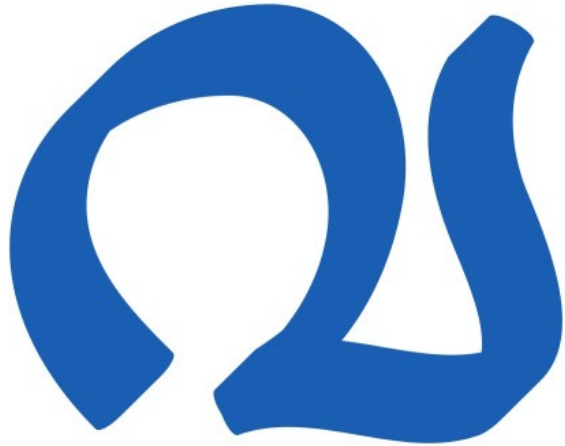

(d) Envelope of Figure 15c

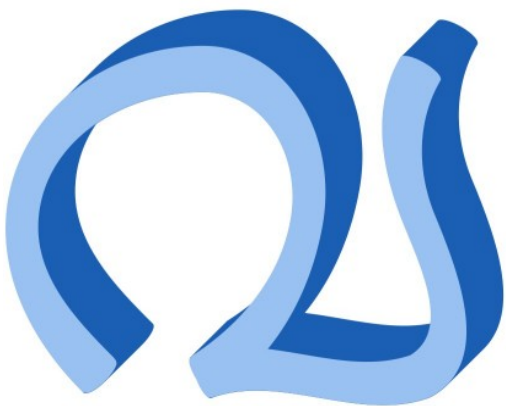

(e) Nupuram regular placed on top of Figure 15d

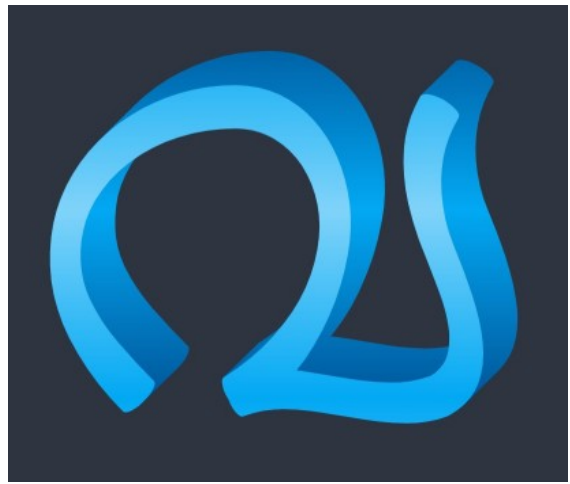

(f) Color gradients applied on Figure 15e. Final rendering.

Figure 15: Construction of Nupuram Color glyph

If we look at 15c carefully, this is a 3D structure, but a hollow one. The facing size and backside is void, creating a hollow structure. To get a solid 3d shape, let us take the envelope of the whole drawing. What I mean by that is, to take the outline of this structure. See Figure 15d

Now it is not hollow. If you look carefully, you will see it is a 3D letter of 'va'. But to help your eyes for the 3D perspective it need colors, or lighting to get the depth perspective. To start with let us place our original 'va' outline on top of it in a lighter color. And Let us make the lighting and coloring a bit more realistic by using color gradients. We get the final rendering as Figure 15f

Nupuram Color font gives 18 predefined pallettes that can be selected by users. Or a user can specify the colors using CSS for example. This color font uses 3 colors for its shadow-ish look. They are Dark, Light, Base colors. Base is the facing color, Light is the central glowing area color. Dark is the color for the shadow part. The colors are used to create a gradient internally.

Nupuram is the first color font in Malayalam.

Each glyph in Nupuram color is constructed by two layers. The facing layer is Nupuram Regular. The background layer is constructed using the techniques explained above. Then each of these layered glyphs are compiles to a UFO. For each layer we define a linear gradient from top to middle and then reflect vertically. By chosing related colors for the gradients in front and background layers, we get the illusion of a 3D object illuminated. This kind of sophisticated glyph manipulation and generation of layers was possible because of `METAPOST`'s easy experimentation and variant generation. Drawing each of these glyphs by hand in a GUI environment would be quite tiresome and would take months of effort.

## 5.8  Nupuram Dots and Arrows

Since a pen could be any shape that can over a predefined path, one experiment I did in Nupuram is to produce a variant where the paths are filled with equally placed dots. We get a dotted font, often used in educational context - to learn how to write a letter by writing on top of it. We call this variant Nupuram Dots. See Figure 16

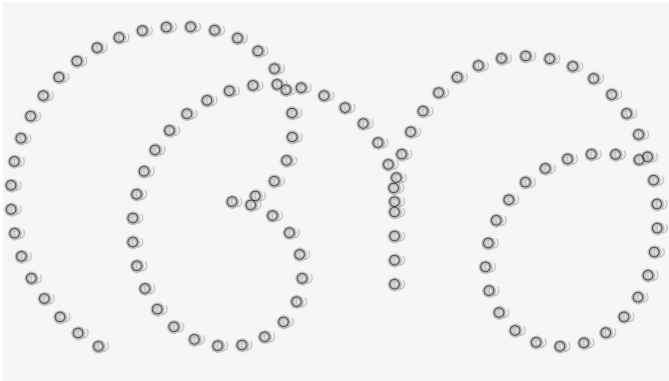

Figure 16: Nupuram Dots - an font for learning to write.

I tried to replace dots with arrows to indicate the writing direction. These arrows are equally placed on the pen path. See Figure 17a. To make it more useful, I placed the arrows font on top of the regular font. And gave two colors for each layer making it a color font. This color can be customized at user side like we explained in the section about Nupuram color. See Figure 17b

Special fonts like dot fonts play a valuable role in facilitating the process of learning to write letters, especially for young learners and those with fine motor skill challenges. By tracing arrows or connecting the dots, students develop muscle memory and hand-eye coordination necessary for proper letter formation. Dot fonts are particularly popular in early childhood education, where they help children transition from drawing shapes to writing letters.

Developing a dots or arrow font like this is a full-blown typeface project in common typeface design approach. Here, we see that it is a quick derivation of an existing typeface.

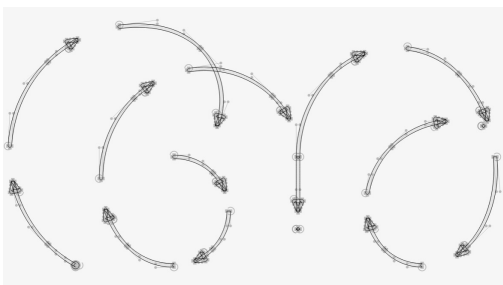

(a) Nupuram Arrows - Showing writing direction

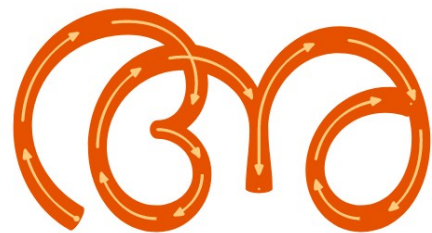

(b) Nupuram arrows color font with default color palette.

Figure 17: Nupuram Arrows

Optimizing these curves with the curve harmonization techniques were required for aesthetic reasons. Linus Romer's implementation of Curvature combs and harmonization [11] of META-

`POST` was used for the purpose.

## 6 Type design workflow

A design and development workflow fine tuned for this kind of type design was also required. It is important to see the results of a `METAPOST` instantly for fast feedback loops. I integrated `METAPOST` based type design to popular Visual Studio Code so that the designer can work on `METAPOST` code and live preview the results with all design utilities like design grid, point coordinates, anchors and so on. I also prepared an online sandbox for `METAPOST` so that quick and cheap experiments are possible, and also sharing and collaboration on design concepts are feasible[18].

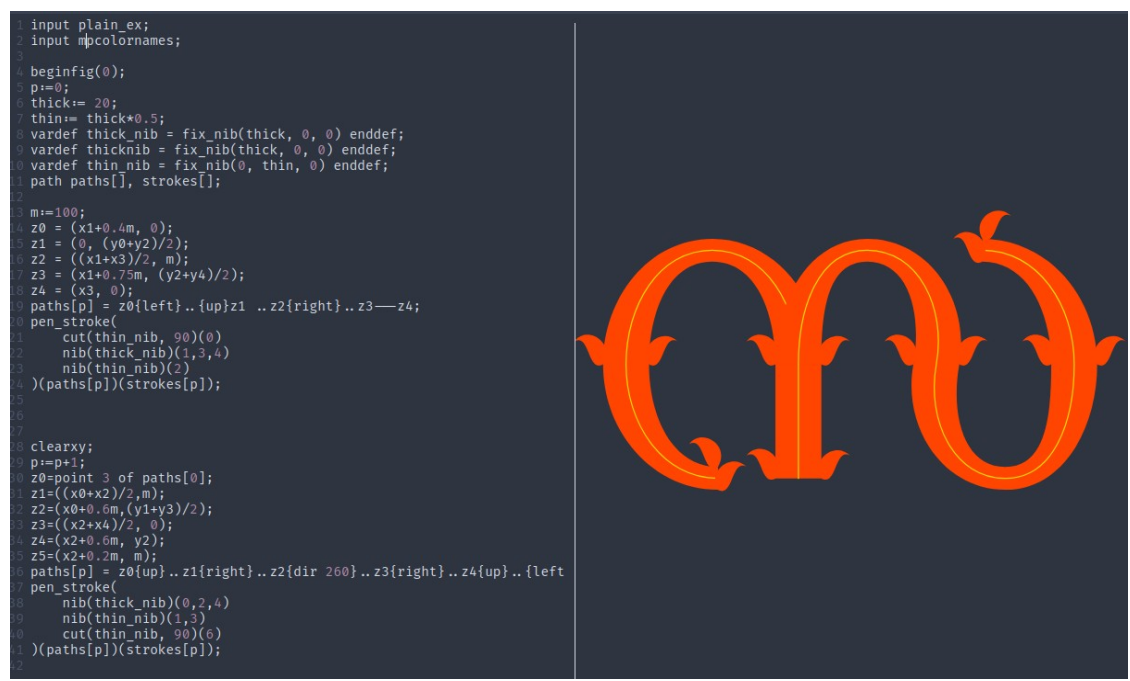

Figure 18: `METAPOST` code and live preview setup. Glyph designed by the author.

## 7 Derivatives

The educational typefaces with arrows and dots were first of this kind in Malayalam language. I used the SVGs produced out of `METAPOST` code to create a website to learn writing Malayalam[19]. The SVG stroke paths are animatable to simulate a drawing animation. Along with

---

[18]The metapost sandbox https://mpost.thottingal.in

[19]https://learn.smc.org.in

Arrows, Dots variants and incorporating example words and pronunciation audio samples, this website has serious users now.

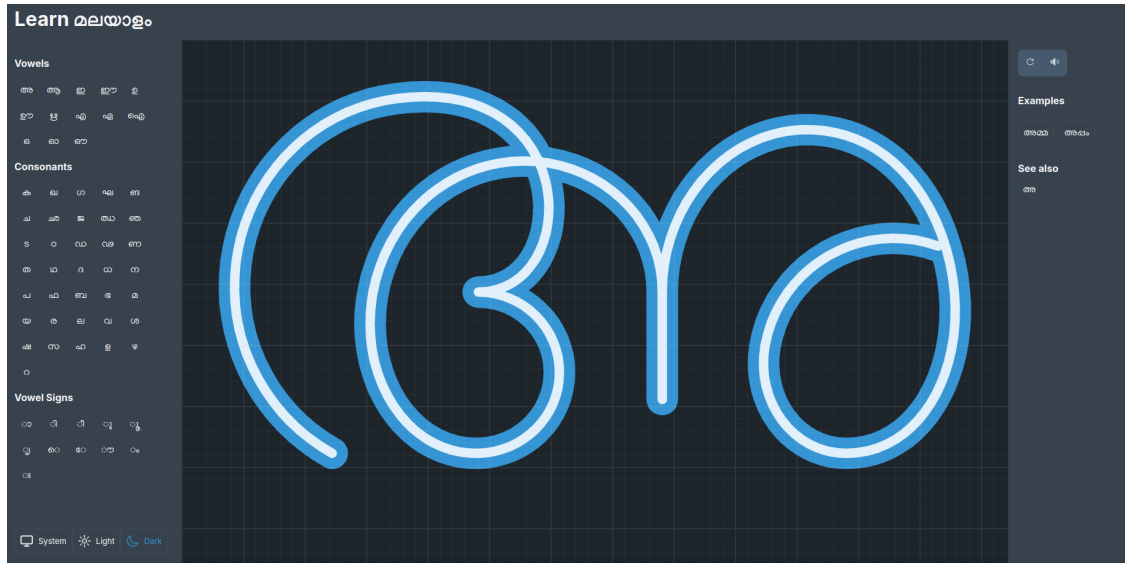

Figure 19: Educational website to learn how to write letters of Malayalam, created using the same SVGs used for the typeface.

# 8 Malini typeface

After Nupuram, I also designed another typeface with the same workflow, a typeface optimized for body text, which has stricter stroke modulation, optical fine tuning requirements. Malini has 4 variable axes: Weight, Width, Slant and Optical Size. Malini is first typeface in Malayalam script with optical sizing. While Nupuram allowed more freedom in drawing as it is designed as display typeface and has handwriting like characteristics, Malini demanded more precise drawing. This challenge helped to fine tune the workflow again and retrospectively apply some of the improvements in Nupuram typeface.

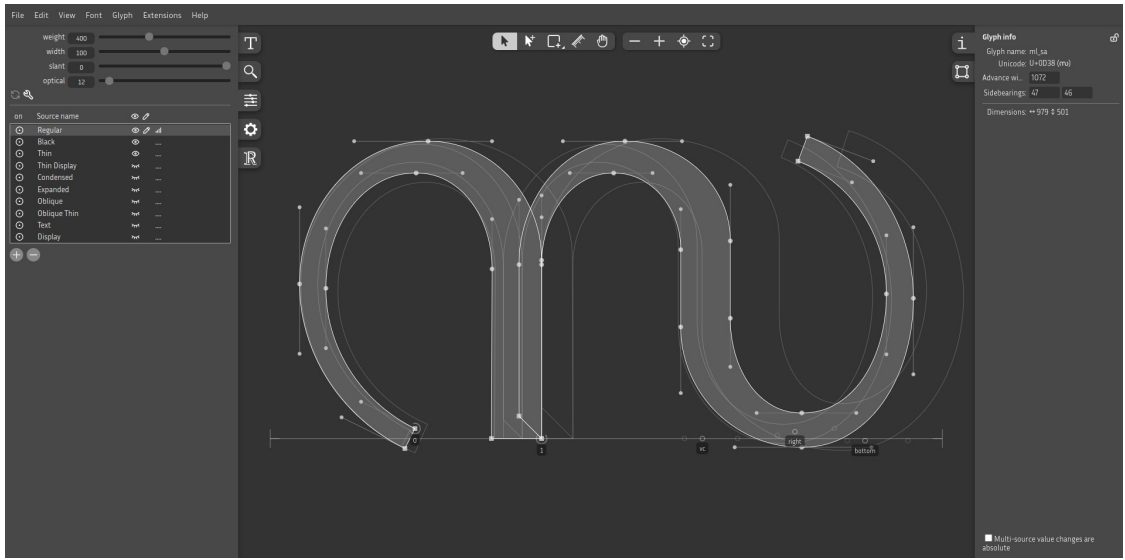

Figure 20: Malini typeface - Letter Sa of Malayalam showing stroke modulation and various axes

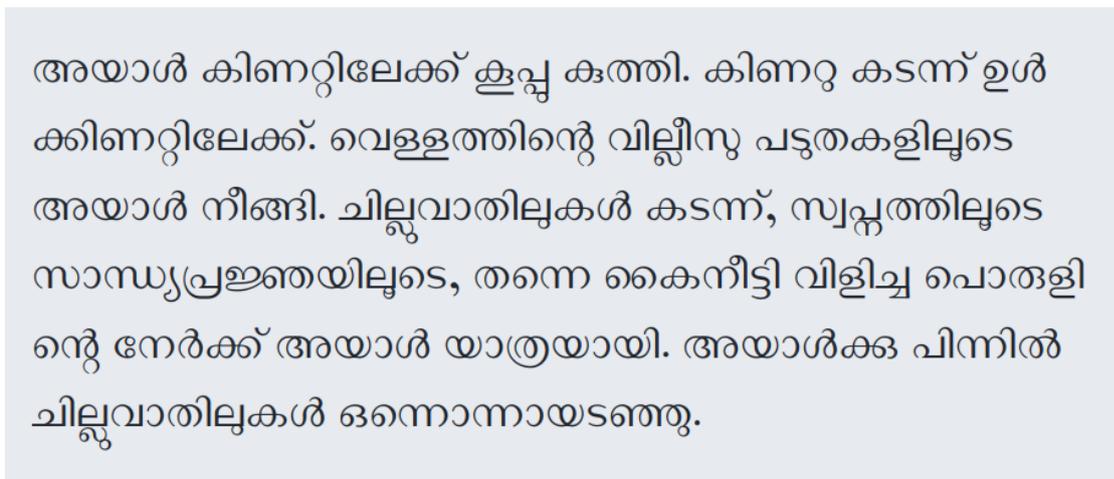

Figure 21: Malini sample text rendering

# 9 Lessons

Parametric approach helped me to do easy experimentation. Traditional type design approach is not flexible once the key typographic metrics are fixed. In METAPOST based approach, fine tuning these parameters is possible at any stage of the project. It is very easy to experiment with different parameters and pick one that is satisfactory to the designer. Sometimes these experiments also reveals surprising results. The approach for 3D like color font was born from such an experiment.

Leveraging my background as a software engineer, adept at working with source code, I found that applying this familiar paradigm to type design significantly enhanced productivity. Shapes were manipulated as modularized methods within the source code, and this modular approach proved instrumental in generating design outputs that were not only consistent, but also coherent. This marked a notable departure from the challenges I had encountered in traditional typeface design projects in the past, where maintaining consistency and cohesiveness had often been a struggle.

Producing a superfamily of Malayalam typeface with multiple variable and color fonts can easily take multiple years of effort, and the collaboration of multiple individuals. However majority of Nupuram's work was finished in my free time in 6 months.

Capturing the details of all required glyphs and their opentype features in program was not an easy task, but that is an investment for future fonts. I could reuse most of the METAPOST code and the framework in Malini typeface project[20] even though that is a typeface optimized for body text.

The ability to define the design precisely in mathematical language will guide the designer for stringent mathematical proportions for letter forms. Nevertheless, it is essential to recognize that such precision does not inherently guarantee the production of aesthetically pleasing glyphs. This tension between mathematical precision and aesthetic appeal emerged as a recurring theme throughout my type design process. For example, the programmer's inclination to align an element exactly at the 0.5 mark of the entire width versus the more subjective choice of proportions that align with visual harmony and appeal.

The ability to swiftly generate variations and, consequently, produce variable fonts proved to be a significant advantage in my work. As an advocate of free and open-source principles, my objective was to construct the entire type design workflow and develop typefaces without relying on proprietary software. Notably, at the time, there was a lack of reliable free software tools for typeface design tailored to the creation of variable fonts. I overcame that limitation with this new workflow. Nupuram is one of the early color fonts after publishing opentype colrv1 specification. It is worth noting that the support for variable and color fonts are yet to catchup in common software application people use.

It is tempting to proliferate numerous subfamilies from a single typeface due to the ease of doing so. In this regard, it is pertinent to recall Knuth's cautionary advice regarding this issue. In the Nupuram project, driven by a desire to explore the myriad possibilities, I did create numerous subfamilies. Nevertheless, it remains a verifiable fact that an abundance of variations does not necessarily equate to an abundance of genuinely useful fonts.

---

[20]Malini is a new typeface project that the author working on with METAPOST based workflow https://github.com/smc/malini

I must acknowledge that METAPOST is not a programming language one can easily learn. It diverges significantly from the conventions of popular programming languages, presenting a distinctive learning curve. While valuable documentation resources do exist, there is a noticeable dearth of practical examples that designers can readily reference and learn from. Given this challenge, prior to advocating for the adoption of this workflow among fellow type designers, I recognized the need to address this limitation. So I documented Nupuram extensively. The project is published in free and open source license for anybody to refer and learn[21]. I created a simple METAPOST playground website https://mpost.thottingal.in where people can quickly write METAPOST code and preview the result. I started a repository of various type design concepts illustrated using METAPOST[22].

Upon embarking on this exploration using METAPOST, I encountered several articles outlining why METAPOST or METAFONT did not catch up. A recurring theme in these discussions was the debate over whether an arbitrary type design can be expressed by the pen based approach where outlines are essentially the outputs of pen strokes. It is undeniable that outline-based design has established itself as a tried-and-true, successful design methodology, and this criticism is indeed grounded in validity. However, I contend that there is no inherent necessity for the exclusive use of pen-based design in every aspect of glyph creation. Whether it manifests as a stroke path or an outline, it essentially boils down to an array of coordinates from my perspective. The manipulation of arrays is a skill that I have honed as a programmer. For example, The serif shape for a Latin letter cannot be created using the outlines of a pen. But nothing prevents a METAPOST programmer to quickly draw that outline without a pen definition.

---

[21]Nupuram source code repository: https://github.com/smc/nupuram. Licensed under SIL Open Font License, Version 1.1

[22]Typeface design concepts illustrated using METAPOST https://github.com/santhoshtr/type-concepts

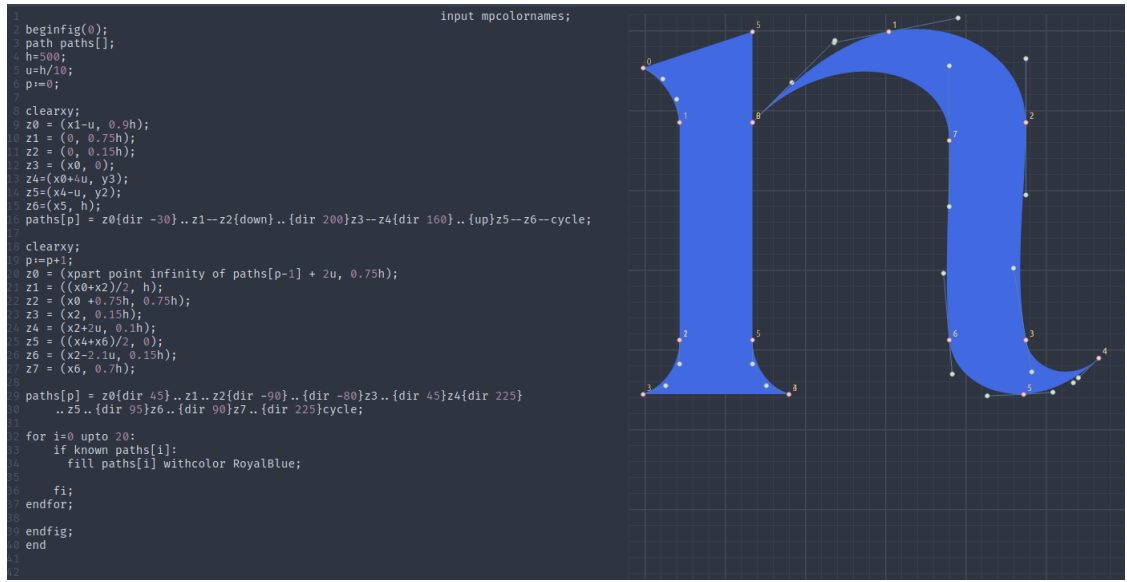

Figure 22: Letter 'n' drawn using METAPOST using outline technique, without using any pens

## 10   Conclusion

In this paper, I have provided an outline of the parametric type design methodology, using the Nupuram typeface as an illustrative example. I also elaborated the lessons learned. While this new approach enabled me to create a typeface with many features hitherto deemed unattainable, it is important to underscore that I am not proposing it as a replacement for the existing type design workflow. The approach outlined here is suitable for designers who possess proficiency in software engineering. As typefaces are becoming sophisticated software these days, I am hopeful that this approach will find utility among a broader spectrum of designers. I aspire to see the Nupuram and Malini typefaces, born from this exploration, become valuable assets for a diverse community of users.